\documentclass[11pt]{article}
\usepackage{nodalida2025}
\usepackage{times}
\usepackage{url}
\usepackage{latexsym}
\usepackage{xcolor}
\usepackage{colortbl}
\usepackage{hyperref}
\usepackage{graphicx}
\usepackage{array}
\usepackage{amsmath}
\usepackage{enumitem}
\usepackage{booktabs}
\usepackage[belowskip=-0.2em,labelfont=bf,labelsep=endash]{caption}
\usepackage{tcolorbox}

\definecolor{primary}{HTML}{9fd4fc}
\definecolor{secondary}{HTML}{CD9ACD}
\definecolor{tertiary}{HTML}{88CCFF}

\definecolor{darkblue}{rgb}{0, 0, 0.5}
\hypersetup{colorlinks=true,citecolor=darkblue,linkcolor=darkblue, urlcolor=darkblue}

\newcommand{\example}[1]{{\vspace{-0.35em}{\centering\begin{tcolorbox}[width=0.98\columnwidth,colback={primary},left=2pt,right=2pt,top=2pt,bottom=2pt,boxrule=0pt, sharp corners]\footnotesize#1\end{tcolorbox}}\vspace{-0.35 em}}}

\aclfinalcopy %

\title{The Roles of English in Evaluating Multilingual Language Models}

\author{Wessel Poelman \and
  Miryam de Lhoneux \\
  Department of Computer Science \\
  KU Leuven, Belgium \\
  {\tt \{wessel.poelman, miryam.delhoneux\}@kuleuven.be}}

\date{}

\begin{document}
\maketitle
\begin{abstract}
Multilingual natural language processing is getting increased attention, with numerous models, benchmarks, and methods being released for many languages.
English is often used in multilingual evaluation to prompt language models (LMs), mainly to overcome the lack of instruction tuning data in other languages.
In this position paper, we lay out two roles of English in multilingual LM evaluations: as an \emph{interface} and as a \emph{natural language}.
We argue that these roles have different goals: \emph{task performance} versus \emph{language understanding}.
This discrepancy is highlighted with examples from datasets and evaluation setups.
Numerous works explicitly use English as an interface to boost task performance.
We recommend to move away from this imprecise method and instead focus on furthering language understanding.
\end{abstract}

\section{Introduction}\label{sec:intro}
With the increase of in-context, prompt-based evaluation of auto-regressive languages models  \cite[LMs,][]{brown2020language}, choices have to be made on how prompts are created.
Specifically in multilingual evaluation, a crucial choice is in which language(s) prompts are written.
In practice, English tends to be mixed with a target language with the explicit goal of increasing \emph{task performance}.
We argue this goal is different from furthering \emph{language understanding}.
In this position paper, we outline two roles of English at the core of this discrepancy and their implications.

Several works have highlighted methodological issues in multilingual evaluation setups \cite{artetxe2020call,ploeger2024what}.
The dominance of English in natural language processing (NLP) has also been discussed repeatedly \cite{joshi2020state,ruder2022squarea}.
With the increase of prompt-based evaluations of models, a new issue has appeared: English being used as an \emph{interface}, rather than a \emph{natural language}.

In recent work, \citet{zhang2023dont} propose a taxonomy of prompt-based multilingual LM evaluations.
They conclude that \emph{``[the model] achieves higher performance when the task is presented in English.''}
This finding is consistent among a large number of papers \cite[inter alia]{shi2022language,huang2022zeroshot,fu2022polyglot,lin2022fewshot,asai2024buffet,etxaniz2024multilingual}.
Resorting to using English like this is hardly surprising given that instruction tuning datasets are expensive to create and not readily available for most languages.
Less surprising still is the finding that English performs well, as it is included in virtually all LMs.
It does bring into question: what is being evaluated and what do we learn from this?

To illustrate: \mbox{MaLa-500} \cite{lin2024mala500b} is a Llama 2-based model \cite{touvron2023llamaa} that underwent continued pre-training in over 500 languages.
It is partially evaluated on a news topic classification task using SIB-200 \cite{adelani2024sib200}, a dataset of \mbox{$(\emph{sentence}, \emph{topic})$}~pairs in 205 languages.
The model is prompted as follows:

\example{The topic of the news \textbf{\{sentence\}} is \textbf{\{topic\}}}

\noindent
Using the prompt with a Turkish\footnote{English translations of examples are in Appendix \ref{app:examples}.} example gives:

\example{The topic of the news Bu oteller günün zenginlerinin ve ünlülerinin kalacağı yerlerdi ve çoğu zaman kaliteli yemeklere ve gece hayatına sahipti. is entertainment}

This format is used across all 205 languages in few-shot setups from one to ten.
This mixture of English and a target language is, arguably, not very `natural'.
We refer to this role of English as an \emph{interface}, rather than a \emph{natural language}.
In the next sections, we outline these roles and why they are important to consider in multilingual evaluation.

\begin{figure*}[ht]
    \centering
    \includegraphics[width=\textwidth]{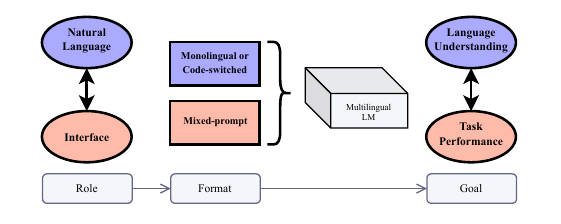}
    \caption{Schematic overview of the different roles of English in multilingual LM evaluation.}
    \label{fig:overview}
\end{figure*}

\section{Evaluation Goals}\label{sec:goals}

\paragraph{Language understanding.}
We take the common perspective that evaluation concerns a \emph{task} which is used as a proxy for \emph{understanding}.
This is exemplified by the \emph{natural language understanding} (NLU) label many datasets and models adhere to (including SIB-200).
A news topic classification task shows that the model (arguably) `understands' some of the differences between news categories.
A model that rewrites, translates or summarizes `understands' both task instructions and target passages.
In a multilingual setting, the understanding of interest is \emph{generalizability} across languages; a model performing a task in a target language supposedly \emph{understands} something about that language.
This is then applied to multiple languages.
We refer to this as `multilingual natural language understanding' (MLU).
Specifically, we use MLU to mean `understanding a target language is part of multilingual natural language understanding.'\footnote{We are aware this (ab)use of terminology is not standard.}

Understanding English by itself and understanding a \emph{natural} mix of English and another language are both part of MLU.
The latter enters the domain of code-switching: the phenomenon where a speaker fluently switches between multiple different languages during the same conversational turn \cite{milroy1995one}.\footnote{Some differentiate between code-switching and code-mixing, we do not make a distinction. For an overview of code-switching in NLP, we refer to \citet{winata2023decades}.}

The MaLa-500 prompt mixes English and a target language.
However, it is hard to classify this as code-switching, as the switch is hardly natural, especially in a few-shot setup.
Rather than a \emph{natural language} that tells something about \emph{language understanding}, English is used as an \emph{interface} to the LM with the goal of increasing \emph{task performance}.
We refer to this mixing as a \emph{mixed-prompt}.

\paragraph{Task performance.}
Another widespread perspective on evaluation in (multilingual) NLP considers performance on a task as an end in itself.\footnote{We thank two reviewers for suggesting to put more emphasis on this perspective.}
If we want to classify news topics in a practical application operating in a multilingual setting, what a model supposedly understands or how well it models a particular language is of little value.
What matters is the system performing its task adequately across languages.
Without using English, the system might not even work at all.
This is a common justification; mixing in English is arguably better than not having a system at all.

While practical, this perspective is seemingly at odds with the many tasks and datasets that present themselves under the aforementioned label of language \emph{understanding}.
Additionally, task performance as the sole goal introduces a usability issue.
Auto-regressive LMs are increasingly meant to be directly interacted with (a \emph{natural} language interface).
If we have to resort to a mixed-prompt for the system to even function, it means the user has to be able to write English and get familiar with this unnatural mixing of languages.

Figure \ref{fig:overview} summarizes our argument and terminology.
Next, we provide more details regarding the discrepancies between using English as an interface versus using it as a natural language.

\section{Evaluation Methods}\label{sec:methods}
As mentioned in \S \ref{sec:intro}, a large body of contemporary research in multilingual NLP focuses on prompting methods.
Common evaluation setups range from (i) prompts fully in a target language, to (ii) English instructions with task-specific passages in the target language, to (iii) translating all text into English before presenting it to a model.\footnote{We do not further discuss `translate everything' as this resembles evaluating English as a \emph{natural language}.}
None of these works refer to this mixture as being code-switched text.
All conclude that a mixture of English and a target language (a mixed-prompt) generally results in the best task performance.
In this section we show why a mixed-prompt is an inherently imprecise method to use in evaluation, even if maximizing task performance is the goal.

If we use a prompt fully in a target language, we are clearly evaluating part of MLU.
A mixed-prompt introduces \emph{additional factors} that are evaluated that are neither the task nor MLU.
We illustrate this from two angles: the representation of the prompt and fortuitous issues from unnaturally mixing English and a target language.

Consider how to evaluate a multilingual masked language model on the news classification task.
A classification layer is added to a pre-trained model to predict the topic labels; it sees label \emph{indices} that are consistent across languages.
The labels are language-agnostic for the model (i.e., detached from natural language).
The evaluation method and goal are clear: mapping a target language sequence to one of these indices.
There are no additional signals influencing this process.

In a prompting setup, the representation of the labels can either be language-agnostic (numbers, letters, symbols, etc.), or not (English words, target language words, etc.).
These options result in any number of \emph{tokens}, which will have different representations within the model, unless specifically accounted for.
In many multilingual evaluation prompts, the classification labels are English words (such as in the MaLa-500 example).
Without target language words or (to an extent) language-agnostic labels, the evaluation method and goal will be inherently imprecise.

In addition to the different representation, more than just the task is evaluated with a mixed-prompt setup.
To illustrate this, consider the following setup from the AfriMMLU subtask of IrokoBench \cite{adelani2024irokobencha}:

\example{
    You are a highly knowledgeable and intelligent
    
    artificial intelligence model answers multiple-choice
    
    questions about \textbf{\{subject\}}
    
    Question: \textbf{\{question\}}
    
    Choices:\\
            A: \textbf{\{choice1\}}\\
            B: \textbf{\{choice2\}}\\
            C: \textbf{\{choice3\}}\\
            D: \textbf{\{choice4\}}\\
    Answer:
}

\noindent
The prompt and \texttt{subject} are always in English, the \texttt{question} and \texttt{choices} in the target language.
With this setup, more is tested than just a task in a target language:

\begin{itemize}[topsep=1pt,parsep=0pt,partopsep=0pt,itemsep=0.5em]
    \item Code-switching, if this is considered natural, or unnatural `mixed-prompt' switching.
    \item Script-switching, if the target language uses a non-Latin script (which applies to Amharic in IrokoBench, using the Ge`ez script).
    \item Instruction following in English.
    \item Grammatical error correction in English.\footnote{We have notified the AfriMMLU authors about this. The typo is in the prompt in the paper and in the \emph{lm-evaluation-harness} \cite{biderman2024lessons}, which is used to obtain their results: {\tiny \url{https://github.com/EleutherAI/lm-evaluation-harness/blob/7882043b4ee1ef9577b829809c2f4970b0bdba91/lm_eval/tasks/afrimmlu/direct/utils.py}}.}
    \item Answering high-school level exam questions in the target language.
\end{itemize}
With these mixed-prompts, we arguably do not test MLU, as that would entail a native target language prompt.
At the same time, we test more than just the task, even though that is the explicit goal of using English in this way.

While we only discussed classification tasks until now, our argument also applies to other types of tasks.
Consider the following zero-shot machine translation prompt from \citet{hendy2023how}:

\example{
Translate this sentence from \textbf{\{source\}} to \textbf{\{target\}}\\
Source: \textbf{\{source\_sentence\}}\\
Target:
}

\noindent
The prompt is always in English, the \texttt{source} and \texttt{target} are English words referring to the languages, and the \texttt{source\_sentence} is in the target language.
Filled in, it looks like this:

\example{
\# DE $\rightarrow$ NL\\
Translate this sentence from German to Dutch\\
Source: Du gehst mir auf den Keks\\
Target:\\

\# NL $\rightarrow$ DE\\
Translate this sentence from Dutch to German\\
Source: tijd voor een bakje koffie\\
Target:
}

\noindent
The authors mention they \emph{``explore prompt selection strategies along two dimensions: quality and relevance''}, but do not mention target language prompts.
To underline the \emph{interface} role of English: it is neither the translation source nor target here.
\citet{hendy2023how} mention that \emph{``keeping the prompt format the same allows us to potentially leverage the benefits of the underlying instruction finetuning protocol to the full extent.''}
This makes explicit the goal of \emph{task performance}.
Prompting a model to translate a sentence is easily done in a manner that more closely aligns with the goal of MLU, does not use English, and is closer to natural code-switching:

\example{
\# DE $\rightarrow$ NL (Dutch speaker)\\
Wat betekent ``Du gehst mir auf den Keks'' in het Nederlands?\\

\# NL $\rightarrow$ DE (Dutch speaker)\\
Hoe zeg je ``tijd voor een bakje koffie'' in het Duits?
}

\section{Why does this matter?}\label{sec:matter}
Interacting with computers in a natural manner is arguably the ultimate goal of numerous sub-fields of computer science.
Work on natural language interfaces to information systems dates back decades \cite{winograd1972understanding,waltz1978english}.
LMs bring us ever closer to this goal.
However, in a multilingual setting, it is important to consider what \emph{natural language} is, what is being evaluated, and what promises are sold.
Next, we outline the implications of the \emph{interface} versus \emph{natural language} roles on evaluation practices.

\paragraph{Interface.}
Let us start with the role in which English is akin to a programming language.\footnote{Also reflected in this famous post: \url{https://x.com/karpathy/status/1617979122625712128}}
We need an interface to communicate with a system, in a way the system can understand.
We have seen that mixed-prompts are used to get the system to perform better on a given task.
Given the scarcity of instruction tuning datasets and the costs involved in creating these, it is understandable that this is a common (albeit sometimes implicit) perspective.
English becomes the `programming' language that glues target language passages together and makes the system perform a task.
Programming languages also predominantly use English labels for their keywords.
However, if the keyword for a \verb|while| loop happens to be \verb|mientras| or \verb|kjsdfk| is irrelevant for its function.
These are natural language-agnostic as the meaning (as interpreted by a compiler or interpreter) does not change.
Variable names and keywords can be chosen arbitrarily.\footnote{Within the specifications of the programming language.}
This is not the case with prompting, which is sensitive to slight changes, both in English \cite{sclar2023quantifying} and multilingual setups \cite{zhang2023dont,asai2024buffet}.

Additionally, evaluation setups that use English as an interface introduce knowledge leakage from English to the target language.
This is, again, with the explicit goal of improving task performance.\footnote{Knowledge leakage also explicitly happens in parameter sharing \cite{zeman2008crosslanguage} or cross-lingual transfer \cite{philippy2023common}.
However, these methods are fundamentally different from mixed-prompts as they (i) treat English as a natural language, and (ii) target knowledge sharing at the training or finetuning phase, not the evaluation phase.}
Being able to understand English instructions is not the same as being able to understand target language instructions.
If English truly was a programming language, this would not matter, as the meaning of the instructions would be separate from the meaning of the target language passages.
Given that English is a natural language, this \emph{de facto} means more is evaluated than just the task.
Consequently, such evaluations are imprecise at best, as shown in \S \ref{sec:methods}.

Prompt-based evaluations should extend MLU to the \emph{instruction} domain.
A mixed-prompt setup claiming to test \mbox{\emph{``multilingual understanding''}} might more accurately be described as \emph{``understanding English instructions interleaved with passages from target language(s), albeit not in a natural code-switching setup.''}

\paragraph{Natural language.}
When we consider the other role of English in multilingual prompt-based evaluation, we should treat it the same as any other language.
The `Multilingual Exemplars' setup from \citet{shi2022language} is a creative interpretation of this perspective.
In this few-shot setup, the model sees various examples, all in \emph{different} languages.
The final question is asked in the target language.
A setup like this extends the definition of `multilingual language understanding' to the extreme.
It becomes harder to interpret what a multilingual model knows about any individual language in this context, but English is certainly not an interface, it is a natural language like all others.

A less extreme setup would simply use native, target language prompts or natural code-switched prompts.
This is costly, but it aligns much better with the goal of multilingual natural language understanding.
Indeed, several works specifically explore this direction \cite{kopf2023openassistant,singh2024aya}.
This approach clearly tests multilingual language understanding, including the instruction domain.
If performance on a particular task in a particular language is lagging behind, or not working at all, it means focus should be put on addressing the core of these issues (e.g., data or modeling).
Ideally, we should not resort to imprecise methods to boost task performance.

\section{Conclusion}
In this position paper we outline two roles of English in multilingual language model evaluation: as an \emph{interface}, with the goal of \emph{task performance}, and as a \emph{natural language}, with the goal of \emph{language understanding}.
We (i) list works that incorporate English with the explicit goal of boosting task performance, even in tasks such as translation where it is neither the source nor target, underlining the \emph{interface} role, (ii) show that mixing English with a target language in a \emph{mixed-prompt} is unnatural (i.e., not code-switching), and (iii) outline why the interface role is an imprecise choice when evaluating multilingual language understanding of language models.

Additionally, we argue that using a mixed-prompt tests \emph{more} than just performance on a certain task.
Because English is a natural language and not a programming language, using it in a mixed prompt will inherently lead to fortuitous factors such as (un)natural switching between languages or scripts, grammatical error correction, and more.
This all results in imprecise or misleading evaluations, even if the ultimate goal was to evaluate and improve task performance.

We finally contrast the implications of the two roles on evaluation practices.
We recommend to move away from using English as an interface in multilingual evaluations and ultimately advocate for the goal of \emph{language understanding}.

\section*{Acknowledgments}
WP is funded by a KU Leuven Bijzonder Onderzoeksfonds C1 project with reference C14/23/096.
We thank the LAGoM-NLP group at KU Leuven for valuable paper recommendations and Mahdi Dhaini for reviewing an early draft of this paper.
We also thank the reviewers for their constructive comments.

\vfill
\newpage

\bibliographystyle{acl_natbib}
\bibliography{nodalida2025}

\begin{thebibliography}{27}
\expandafter\ifx\csname natexlab\endcsname\relax\def\natexlab#1{#1}\fi

\bibitem[{Adelani et~al.(2024{\natexlab{a}})Adelani, Liu, Shen, Vassilyev, Alabi, Mao, Gao, and Lee}]{adelani2024sib200}
David Adelani, Hannah Liu, Xiaoyu Shen, Nikita Vassilyev, Jesujoba Alabi, Yanke Mao, Haonan Gao, and En-Shiun Lee. 2024{\natexlab{a}}.
\newblock \href {https://aclanthology.org/2024.eacl-long.14} {{{SIB-200}}: {{A Simple}}, {{Inclusive}}, and {{Big Evaluation Dataset}} for {{Topic Classification}} in 200+ {{Languages}} and {{Dialects}}}.
\newblock In \emph{Proceedings of the 18th {{Conference}} of the {{European Chapter}} of the {{Association}} for {{Computational Linguistics}} ({{Volume}} 1: {{Long Papers}})}, pages 226--245.

\bibitem[{Adelani et~al.(2024{\natexlab{b}})Adelani, Ojo, Azime, Zhuang, Alabi, He, Ochieng, Hooker, Bukula, Lee, Chukwuneke, Buzaaba, Sibanda, Kalipe, Mukiibi, Kabongo, Yuehgoh, Setaka, Ndolela, Odu, Mabuya, Muhammad, Osei, Samb, Guge, and Stenetorp}]{adelani2024irokobencha}
David~Ifeoluwa Adelani, Jessica Ojo, Israel~Abebe Azime, Jian~Yun Zhuang, Jesujoba~O. Alabi, Xuanli He, Millicent Ochieng, Sara Hooker, Andiswa Bukula, En-Shiun~Annie Lee, Chiamaka Chukwuneke, Happy Buzaaba, Blessing Sibanda, Godson Kalipe, Jonathan Mukiibi, Salomon Kabongo, Foutse Yuehgoh, Mmasibidi Setaka, Lolwethu Ndolela, Nkiruka Odu, Rooweither Mabuya, Shamsuddeen~Hassan Muhammad, Salomey Osei, Sokhar Samb, Tadesse~Kebede Guge, and Pontus Stenetorp. 2024{\natexlab{b}}.
\newblock \href {http://arxiv.org/abs/2406.03368v1} {{{IrokoBench}}: {{A New Benchmark}} for {{African Languages}} in the {{Age}} of {{Large Language Models}}}.
\newblock \emph{arXiv preprint, arXiv:2406.03368v1}.

\bibitem[{Artetxe et~al.(2020)Artetxe, Ruder, Yogatama, Labaka, and Agirre}]{artetxe2020call}
Mikel Artetxe, Sebastian Ruder, Dani Yogatama, Gorka Labaka, and Eneko Agirre. 2020.
\newblock \href {https://aclanthology.org/2020.acl-main.658} {A {{Call}} for {{More Rigor}} in {{Unsupervised Cross-lingual Learning}}}.
\newblock In \emph{Proceedings of the 58th {{Annual Meeting}} of the {{Association}} for {{Computational Linguistics}}}, pages 7375--7388.

\bibitem[{Asai et~al.(2024)Asai, Kudugunta, Yu, Blevins, Gonen, Reid, Tsvetkov, Ruder, and Hajishirzi}]{asai2024buffet}
Akari Asai, Sneha Kudugunta, Xinyan Yu, Terra Blevins, Hila Gonen, Machel Reid, Yulia Tsvetkov, Sebastian Ruder, and Hannaneh Hajishirzi. 2024.
\newblock \href {https://aclanthology.org/2024.naacl-long.100} {{{BUFFET}}: {{Benchmarking Large Language Models}} for {{Few-shot Cross-lingual Transfer}}}.
\newblock In \emph{Proceedings of the 2024 {{Conference}} of the {{North American Chapter}} of the {{Association}} for {{Computational Linguistics}}: {{Human Language Technologies}} ({{Volume}} 1: {{Long Papers}})}, pages 1771--1800.

\bibitem[{Biderman et~al.(2024)Biderman, Schoelkopf, Sutawika, Gao, Tow, Abbasi, Aji, Ammanamanchi, Black, Clive, DiPofi, Etxaniz, Fattori, Forde, Foster, Hsu, Jaiswal, Lee, Li, Lovering, Muennighoff, Pavlick, Phang, Skowron, Tan, Tang, Wang, Winata, Yvon, and Zou}]{biderman2024lessons}
Stella Biderman, Hailey Schoelkopf, Lintang Sutawika, Leo Gao, Jonathan Tow, Baber Abbasi, Alham~Fikri Aji, Pawan~Sasanka Ammanamanchi, Sidney Black, Jordan Clive, Anthony DiPofi, Julen Etxaniz, Benjamin Fattori, Jessica~Zosa Forde, Charles Foster, Jeffrey Hsu, Mimansa Jaiswal, Wilson~Y. Lee, Haonan Li, Charles Lovering, Niklas Muennighoff, Ellie Pavlick, Jason Phang, Aviya Skowron, Samson Tan, Xiangru Tang, Kevin~A. Wang, Genta~Indra Winata, Fran{\c c}ois Yvon, and Andy Zou. 2024.
\newblock \href {http://arxiv.org/abs/2405.14782} {Lessons from the {{Trenches}} on {{Reproducible Evaluation}} of {{Language Models}}}.
\newblock \emph{arXiv preprint, arXiv.2405.14782v2}.

\bibitem[{Brown et~al.(2020)Brown, Mann, Ryder, Subbiah, Kaplan, Dhariwal, Neelakantan, Shyam, Sastry, Askell, Agarwal, {Herbert-Voss}, Krueger, Henighan, Child, Ramesh, Ziegler, Wu, Winter, Hesse, Chen, Sigler, Litwin, Gray, Chess, Clark, Berner, McCandlish, Radford, Sutskever, and Amodei}]{brown2020language}
Tom Brown, Benjamin Mann, Nick Ryder, Melanie Subbiah, Jared~D Kaplan, Prafulla Dhariwal, Arvind Neelakantan, Pranav Shyam, Girish Sastry, Amanda Askell, Sandhini Agarwal, Ariel {Herbert-Voss}, Gretchen Krueger, Tom Henighan, Rewon Child, Aditya Ramesh, Daniel Ziegler, Jeffrey Wu, Clemens Winter, Chris Hesse, Mark Chen, Eric Sigler, Mateusz Litwin, Scott Gray, Benjamin Chess, Jack Clark, Christopher Berner, Sam McCandlish, Alec Radford, Ilya Sutskever, and Dario Amodei. 2020.
\newblock \href {https://proceedings.neurips.cc/paper/2020/hash/1457c0d6bfcb4967418bfb8ac142f64a-Abstract.html} {Language {{Models}} are {{Few-Shot Learners}}}.
\newblock In \emph{Advances in {{Neural Information Processing Systems}}}, volume~33, pages 1877--1901.

\bibitem[{Etxaniz et~al.(2024)Etxaniz, Azkune, Soroa, Lacalle, and Artetxe}]{etxaniz2024multilingual}
Julen Etxaniz, Gorka Azkune, Aitor Soroa, Oier Lacalle, and Mikel Artetxe. 2024.
\newblock \href {https://aclanthology.org/2024.naacl-short.46} {Do {{Multilingual Language Models Think Better}} in {{English}}?}
\newblock In \emph{Proceedings of the 2024 {{Conference}} of the {{North American Chapter}} of the {{Association}} for {{Computational Linguistics}}: {{Human Language Technologies}} ({{Volume}} 2: {{Short Papers}})}, pages 550--564.

\bibitem[{Fu et~al.(2022)Fu, Ng, and Liu}]{fu2022polyglot}
Jinlan Fu, See-Kiong Ng, and Pengfei Liu. 2022.
\newblock \href {https://aclanthology.org/2022.emnlp-main.674} {Polyglot {{Prompt}}: {{Multilingual Multitask Prompt Training}}}.
\newblock In \emph{Proceedings of the 2022 {{Conference}} on {{Empirical Methods}} in {{Natural Language Processing}}}, pages 9919--9935.

\bibitem[{Hendy et~al.(2023)Hendy, Abdelrehim, Sharaf, Raunak, Gabr, Matsushita, Kim, Afify, and Awadalla}]{hendy2023how}
Amr Hendy, Mohamed Abdelrehim, Amr Sharaf, Vikas Raunak, Mohamed Gabr, Hitokazu Matsushita, Young~Jin Kim, Mohamed Afify, and Hany~Hassan Awadalla. 2023.
\newblock \href {http://arxiv.org/abs/2302.09210v1} {How {{Good Are GPT Models}} at {{Machine Translation}}? {{A Comprehensive Evaluation}}}.
\newblock \emph{arXiv preprint, arXiv:2302.09210v1}.

\bibitem[{Huang et~al.(2022)Huang, Ma, Zhang, Wei, and Wang}]{huang2022zeroshot}
Lianzhe Huang, Shuming Ma, Dongdong Zhang, Furu Wei, and Houfeng Wang. 2022.
\newblock \href {https://aclanthology.org/2022.emnlp-main.790} {Zero-shot {{Cross-lingual Transfer}} of {{Prompt-based Tuning}} with a {{Unified Multilingual Prompt}}}.
\newblock In \emph{Proceedings of the 2022 {{Conference}} on {{Empirical Methods}} in {{Natural Language Processing}}}, pages 11488--11497.

\bibitem[{Joshi et~al.(2020)Joshi, Santy, Budhiraja, Bali, and Choudhury}]{joshi2020state}
Pratik Joshi, Sebastin Santy, Amar Budhiraja, Kalika Bali, and Monojit Choudhury. 2020.
\newblock \href {https://aclanthology.org/2020.acl-main.560} {The {{State}} and {{Fate}} of {{Linguistic Diversity}} and {{Inclusion}} in the {{NLP World}}}.
\newblock In \emph{Proceedings of the 58th {{Annual Meeting}} of the {{Association}} for {{Computational Linguistics}}}, pages 6282--6293.

\bibitem[{K{\"o}pf et~al.(2023)K{\"o}pf, Kilcher, von R{\"u}tte, Anagnostidis, Tam, Stevens, Barhoum, Nguyen, Stanley, Nagyfi, Es, Suri, Glushkov, Dantuluri, Maguire, Schuhmann, Nguyen, and Mattick}]{kopf2023openassistant}
Andreas K{\"o}pf, Yannic Kilcher, Dimitri von R{\"u}tte, Sotiris Anagnostidis, Zhi~Rui Tam, Keith Stevens, Abdullah Barhoum, Duc~Minh Nguyen, Oliver Stanley, Rich{\'a}rd Nagyfi, Shahul Es, Sameer Suri, David~Alexandrovich Glushkov, Arnav~Varma Dantuluri, Andrew Maguire, Christoph Schuhmann, Huu Nguyen, and Alexander~Julian Mattick. 2023.
\newblock \href {https://openreview.net/forum?id=VSJotgbPHF&noteId=7UjKcx6tFl} {{{OpenAssistant Conversations}} - {{Democratizing Large Language Model Alignment}}}.
\newblock In \emph{Thirty-Seventh {{Conference}} on {{Neural Information Processing Systems Datasets}} and {{Benchmarks Track}}}.

\bibitem[{Lin et~al.(2024)Lin, Ji, Tiedemann, Martins, and Sch{\"u}tze}]{lin2024mala500b}
Peiqin Lin, Shaoxiong Ji, J{\"o}rg Tiedemann, Andr{\'e} F.~T. Martins, and Hinrich Sch{\"u}tze. 2024.
\newblock \href {http://arxiv.org/abs/2401.13303v2} {{{MaLA-500}}: {{Massive Language Adaptation}} of {{Large Language Models}}}.
\newblock \emph{arXiv preprint, arXiv:2401.13303v2}.

\bibitem[{Lin et~al.(2022)Lin, Mihaylov, Artetxe, Wang, Chen, Simig, Ott, Goyal, Bhosale, Du, Pasunuru, Shleifer, Koura, Chaudhary, O'Horo, Wang, Zettlemoyer, Kozareva, Diab, Stoyanov, and Li}]{lin2022fewshot}
Xi~Victoria Lin, Todor Mihaylov, Mikel Artetxe, Tianlu Wang, Shuohui Chen, Daniel Simig, Myle Ott, Naman Goyal, Shruti Bhosale, Jingfei Du, Ramakanth Pasunuru, Sam Shleifer, Punit~Singh Koura, Vishrav Chaudhary, Brian O'Horo, Jeff Wang, Luke Zettlemoyer, Zornitsa Kozareva, Mona Diab, Veselin Stoyanov, and Xian Li. 2022.
\newblock \href {https://aclanthology.org/2022.emnlp-main.616} {Few-shot {{Learning}} with {{Multilingual Generative Language Models}}}.
\newblock In \emph{Proceedings of the 2022 {{Conference}} on {{Empirical Methods}} in {{Natural Language Processing}}}, pages 9019--9052.

\bibitem[{Milroy and Muysken(1995)}]{milroy1995one}
Lesley Milroy and Pieter Muysken, editors. 1995.
\newblock \href {https://www.cambridge.org/core/books/one-speaker-two-languages/864B8AA6972F95CB5603629264CF8324} {\emph{One {{Speaker}}, {{Two Languages}}: {{Cross-Disciplinary Perspectives}} on {{Code-Switching}}}}.
\newblock Cambridge University Press.

\bibitem[{Philippy et~al.(2023)Philippy, Guo, and Haddadan}]{philippy2023common}
Fred Philippy, Siwen Guo, and Shohreh Haddadan. 2023.
\newblock \href {https://aclanthology.org/2023.acl-long.323} {Towards a {{Common Understanding}} of {{Contributing Factors}} for {{Cross-Lingual Transfer}} in {{Multilingual Language Models}}: {{A Review}}}.
\newblock In \emph{Proceedings of the 61st {{Annual Meeting}} of the {{Association}} for {{Computational Linguistics}} ({{Volume}} 1: {{Long Papers}})}, pages 5877--5891.

\bibitem[{Ploeger et~al.(2024)Ploeger, Poelman, {de Lhoneux}, and Bjerva}]{ploeger2024what}
Esther Ploeger, Wessel Poelman, Miryam {de Lhoneux}, and Johannes Bjerva. 2024.
\newblock \href {https://aclanthology.org/2024.emnlp-main.326} {What is ``{{Typological Diversity}}'' in {{NLP}}?}
\newblock In \emph{Proceedings of the 2024 {{Conference}} on {{Empirical Methods}} in {{Natural Language Processing}}}, pages 5681--5700.

\bibitem[{Ruder et~al.(2022)Ruder, Vuli{\'c}, and S{\o}gaard}]{ruder2022squarea}
Sebastian Ruder, Ivan Vuli{\'c}, and Anders S{\o}gaard. 2022.
\newblock \href {https://aclanthology.org/2022.findings-acl.184} {Square {{One Bias}} in {{NLP}}: {{Towards}} a {{Multi-Dimensional Exploration}} of the {{Research Manifold}}}.
\newblock In \emph{Findings of the {{Association}} for {{Computational Linguistics}}: {{ACL}} 2022}, pages 2340--2354.

\bibitem[{Sclar et~al.(2023)Sclar, Choi, Tsvetkov, and Suhr}]{sclar2023quantifying}
Melanie Sclar, Yejin Choi, Yulia Tsvetkov, and Alane Suhr. 2023.
\newblock \href {https://openreview.net/forum?id=RIu5lyNXjT} {Quantifying {{Language Models}}' {{Sensitivity}} to {{Spurious Features}} in {{Prompt Design}} or: {{How I}} learned to start worrying about prompt formatting}.
\newblock In \emph{The {{Twelfth International Conference}} on {{Learning Representations}}}.

\bibitem[{Shi et~al.(2022)Shi, Suzgun, Freitag, Wang, Srivats, Vosoughi, Chung, Tay, Ruder, Zhou, Das, and Wei}]{shi2022language}
Freda Shi, Mirac Suzgun, Markus Freitag, Xuezhi Wang, Suraj Srivats, Soroush Vosoughi, Hyung~Won Chung, Yi~Tay, Sebastian Ruder, Denny Zhou, Dipanjan Das, and Jason Wei. 2022.
\newblock \href {https://openreview.net/forum?id=fR3wGCk-IXp} {Language models are multilingual chain-of-thought reasoners}.
\newblock In \emph{The {{Eleventh International Conference}} on {{Learning Representations}}}.

\bibitem[{Singh et~al.(2024)Singh, Vargus, D'souza, Karlsson, Mahendiran, Ko, Shandilya, Patel, Mataciunas, O'Mahony, Zhang, Hettiarachchi, Wilson, Machado, Moura, Krzemi{\'n}ski, Fadaei, Ergun, Okoh, Alaagib, Mudannayake, Alyafeai, Chien, Ruder, Guthikonda, Alghamdi, Gehrmann, Muennighoff, Bartolo, Kreutzer, {\"U}st{\"u}n, Fadaee, and Hooker}]{singh2024aya}
Shivalika Singh, Freddie Vargus, Daniel D'souza, B{\"o}rje Karlsson, Abinaya Mahendiran, Wei-Yin Ko, Herumb Shandilya, Jay Patel, Deividas Mataciunas, Laura O'Mahony, Mike Zhang, Ramith Hettiarachchi, Joseph Wilson, Marina Machado, Luisa Moura, Dominik Krzemi{\'n}ski, Hakimeh Fadaei, Irem Ergun, Ifeoma Okoh, Aisha Alaagib, Oshan Mudannayake, Zaid Alyafeai, Vu~Chien, Sebastian Ruder, Surya Guthikonda, Emad Alghamdi, Sebastian Gehrmann, Niklas Muennighoff, Max Bartolo, Julia Kreutzer, Ahmet {\"U}st{\"u}n, Marzieh Fadaee, and Sara Hooker. 2024.
\newblock \href {https://aclanthology.org/2024.acl-long.620} {Aya {{Dataset}}: {{An Open-Access Collection}} for {{Multilingual Instruction Tuning}}}.
\newblock In \emph{Proceedings of the 62nd {{Annual Meeting}} of the {{Association}} for {{Computational Linguistics}} ({{Volume}} 1: {{Long Papers}})}, pages 11521--11567.

\bibitem[{Touvron et~al.(2023)Touvron, Martin, Stone, Albert, Almahairi, Babaei, Bashlykov, Batra, Bhargava, Bhosale, Bikel, Blecher, Ferrer, Chen, Cucurull, Esiobu, Fernandes, Fu, Fu, Fuller, Gao, Goswami, Goyal, Hartshorn, Hosseini, Hou, Inan, Kardas, Kerkez, Khabsa, Kloumann, Korenev, Koura, Lachaux, Lavril, Lee, Liskovich, Lu, Mao, Martinet, Mihaylov, Mishra, Molybog, Nie, Poulton, Reizenstein, Rungta, Saladi, Schelten, Silva, Smith, Subramanian, Tan, Tang, Taylor, Williams, Kuan, Xu, Yan, Zarov, Zhang, Fan, Kambadur, Narang, Rodriguez, Stojnic, Edunov, and Scialom}]{touvron2023llamaa}
Hugo Touvron, Louis Martin, Kevin Stone, Peter Albert, Amjad Almahairi, Yasmine Babaei, Nikolay Bashlykov, Soumya Batra, Prajjwal Bhargava, Shruti Bhosale, Dan Bikel, Lukas Blecher, Cristian~Canton Ferrer, Moya Chen, Guillem Cucurull, David Esiobu, Jude Fernandes, Jeremy Fu, Wenyin Fu, Brian Fuller, Cynthia Gao, Vedanuj Goswami, Naman Goyal, Anthony Hartshorn, Saghar Hosseini, Rui Hou, Hakan Inan, Marcin Kardas, Viktor Kerkez, Madian Khabsa, Isabel Kloumann, Artem Korenev, Punit~Singh Koura, Marie-Anne Lachaux, Thibaut Lavril, Jenya Lee, Diana Liskovich, Yinghai Lu, Yuning Mao, Xavier Martinet, Todor Mihaylov, Pushkar Mishra, Igor Molybog, Yixin Nie, Andrew Poulton, Jeremy Reizenstein, Rashi Rungta, Kalyan Saladi, Alan Schelten, Ruan Silva, Eric~Michael Smith, Ranjan Subramanian, Xiaoqing~Ellen Tan, Binh Tang, Ross Taylor, Adina Williams, Jian~Xiang Kuan, Puxin Xu, Zheng Yan, Iliyan Zarov, Yuchen Zhang, Angela Fan, Melanie Kambadur, Sharan Narang, Aurelien Rodriguez, Robert Stojnic, Sergey Edunov, and Thomas Scialom. 2023.
\newblock \href {http://arxiv.org/abs/2307.09288v2} {Llama 2: {{Open Foundation}} and {{Fine-Tuned Chat Models}}}.
\newblock \emph{arXiv preprint, arXiv:2307.09288v2}.

\bibitem[{Waltz(1978)}]{waltz1978english}
David~L. Waltz. 1978.
\newblock \href {https://dl.acm.org/doi/10.1145/359545.359550} {An {{English}} language question answering system for a large relational database}.
\newblock \emph{Communications of the ACM}, 21(7):526--539.

\bibitem[{Winata et~al.(2023)Winata, Aji, Yong, and Solorio}]{winata2023decades}
Genta Winata, Alham~Fikri Aji, Zheng~Xin Yong, and Thamar Solorio. 2023.
\newblock \href {https://aclanthology.org/2023.findings-acl.185} {The {{Decades Progress}} on {{Code-Switching Research}} in {{NLP}}: {{A Systematic Survey}} on {{Trends}} and {{Challenges}}}.
\newblock In \emph{Findings of the {{Association}} for {{Computational Linguistics}}: {{ACL}} 2023}, pages 2936--2978.

\bibitem[{Winograd(1972)}]{winograd1972understanding}
Terry Winograd. 1972.
\newblock \href {https://www.sciencedirect.com/science/article/pii/0010028572900023} {Understanding natural language}.
\newblock \emph{Cognitive Psychology}, 3(1):1--191.

\bibitem[{Zeman and Resnik(2008)}]{zeman2008crosslanguage}
Daniel Zeman and Philip Resnik. 2008.
\newblock \href {https://aclanthology.org/I08-3008} {Cross-{{Language Parser Adaptation}} between {{Related Languages}}}.
\newblock In \emph{Proceedings of the {{IJCNLP-08 Workshop}} on {{NLP}} for {{Less Privileged Languages}}}.

\bibitem[{Zhang et~al.(2023)Zhang, Li, Hauer, Shi, and Kondrak}]{zhang2023dont}
Xiang Zhang, Senyu Li, Bradley Hauer, Ning Shi, and Grzegorz Kondrak. 2023.
\newblock \href {https://aclanthology.org/2023.emnlp-main.491} {Don't {{Trust ChatGPT}} when your {{Question}} is not in {{English}}: {{A Study}} of {{Multilingual Abilities}} and {{Types}} of {{LLMs}}}.
\newblock In \emph{Proceedings of the 2023 {{Conference}} on {{Empirical Methods}} in {{Natural Language Processing}}}, pages 7915--7927.

\end{thebibliography}

\newpage
\appendix
\section{Examples}\label{app:examples}
The examples containing Turkish, Dutch or German are repeated here with English translations.

SIB-200 (sample 755):

\example{
The topic of the news Bu oteller günün zenginlerinin ve ünlülerinin kalacağı yerlerdi ve çoğu zaman kaliteli yemeklere ve gece hayatına sahipti. is entertainment\\

The topic of the news \emph{These hotels were where the rich and the famous of the day would stay, and often had fine dining and nightlife.} is entertainment
}

Interface translation examples:

\example{
\# DE $\rightarrow$ NL\\
Translate this sentence from German to Dutch\\
Source: Du gehst mir auf den Keks\\
Target:\\

\# DE $\rightarrow$ NL\\
Translate this sentence from German to Dutch\\
Source: \emph{You're getting on my nerves}\\
Target:
}

\example{
\# NL $\rightarrow$ DE\\
Translate this sentence from Dutch to German\\
Source: tijd voor een bakje koffie\\
Target:\\

\# NL $\rightarrow$ DE\\
Translate this sentence from Dutch to German\\
Source: \emph{time for a cup of coffee}\\
Target:
}

Natural translation examples:

\example{
\# DE $\rightarrow$ NL (Dutch speaker)\\
Wat betekent ``Du gehst mir auf den Keks'' in het Nederlands?\\

\# DE $\rightarrow$ NL (Dutch speaker)\\
\emph{What does} ``Du gehst mir auf den Keks'' \emph{mean in Dutch?}
}

\example{
\# NL $\rightarrow$ DE (Dutch speaker)\\
Hoe zeg je ``tijd voor een bakje koffie'' in het Duits?\\

\# NL $\rightarrow$ DE (Dutch speaker)\\
\emph{How would one say} ``tijd voor een bakje koffie'' \emph{in German?}
}

\end{document}